\documentclass[final,5p,times,twocolumn]{elsarticle}

\usepackage{amssymb}
\usepackage{amsmath}
\DeclareMathOperator{\rank}{rank}

\usepackage{tabularx,ragged2e}
\usepackage{booktabs}
\usepackage{url}
\newcolumntype{Y}{>{\raggedleft\arraybackslash}X}
\usepackage{adjustbox}
\usepackage{color}
\definecolor{myred}{rgb}{.8,.0,.0}

\DeclareMathOperator*{\argmax}{arg\,max}

\begin{document}

\begin{frontmatter}



\title{On dataset transferability in medical image classification}

\author[label1]{Dovile Juodelyte}
\author[label2]{Enzo Ferrante}
\author[label1]{Yucheng Lu}
\author[label3]{Prabhant Singh}
\author[label3]{Joaquin Vanschoren}
\author[label1]{Veronika Cheplygina}

\affiliation[label1]{organization={IT University of Copenhagen},
            addressline={Rued Langgaards Vej 7},
            city={Copenhagen},
             postcode={DK-2300},
            country={Denmark}}
\affiliation[label2]{organization={CONICET - Universidad de Buenos Aires},
            addressline={Pabellon Cero+Infinito – Ciudad Universitaria},
             postcode={(C1428EGA) Ciudad Autonoma de Buenos Aires},
            country={Argentina}}

\affiliation[label3]{organization={Eindhoven University of Technology},
             addressline={5612 AZ},
            city={Eindhoven},
            country={Netherlands}}



\begin{abstract}
Current transferability estimation methods designed for natural image datasets are often suboptimal in medical image classification. These methods primarily focus on estimating the suitability of pre-trained source model features for a target dataset, which can lead to unrealistic predictions, such as suggesting that the target dataset is the best source for itself. To address this, we propose a novel transferability metric that combines feature quality with gradients to evaluate both the suitability and adaptability of source model features for target tasks. We evaluate our approach in two new scenarios: source dataset transferability for medical image classification and cross-domain transferability. 
Our results show that our method outperforms existing transferability metrics in both settings. We also provide insight into the factors influencing transfer performance in medical image classification, as well as the dynamics of cross-domain transfer from natural to medical images. Additionally, we provide ground-truth transfer performance benchmarking results to encourage further research into transferability estimation for medical image classification. Our code and experiments are available at \url{https://github.com/DovileDo/transferability-in-medical-imaging}.
\end{abstract}



\begin{keyword}
Transfer learning \sep Medical imaging \sep Dataset similarity \sep Transferability


\end{keyword}

\end{frontmatter}



\section{Introduction} \label{sec:intro}

Transfer learning has become a cornerstone in medical imaging, offering a solution to the challenge of training deep learning models on limited datasets. By leveraging knowledge from pre-trained models, transfer learning has proven effective in a variety of medical imaging applications \cite{cheplygina2019cats}. A common approach in medical image classification is to pre-train models on ImageNet \cite{deng2009imagenet}, a dataset originally designed for natural image classification. However, unlike natural images, which typically contain distinct global objects, medical images often rely on subtle local texture variations to indicate pathology. Therefore, ImageNet may not always be the optimal source for medical image classification tasks, particularly when working with small datasets \cite{raghu2019transfusion}, where transfer learning is most beneficial.

Prior studies have found that source and target domains should be similar for effective transfer learning \cite{mensink2021factors}. They have shown that pre-training on smaller, closely related source datasets often yields better results on target tasks than using larger but less related source datasets \cite{mensink2021factors,cui2018large}, and that optimal transfer performance is achieved when the source dataset includes images that align with the domain of the target dataset \cite{mensink2021factors}. Furthermore, models pre-trained on ImageNet have demonstrated limitations when applied to medical imaging tasks. They are prone to shortcut learning, where the model relies on spurious correlations rather than learning meaningful representations for medical data \cite{juodelyte2024source,lu2024exploring}, and to memorization \cite{gichoya2022ai}.
\begin{figure}[h!]
    \centering
    \includegraphics[width=0.5\textwidth]{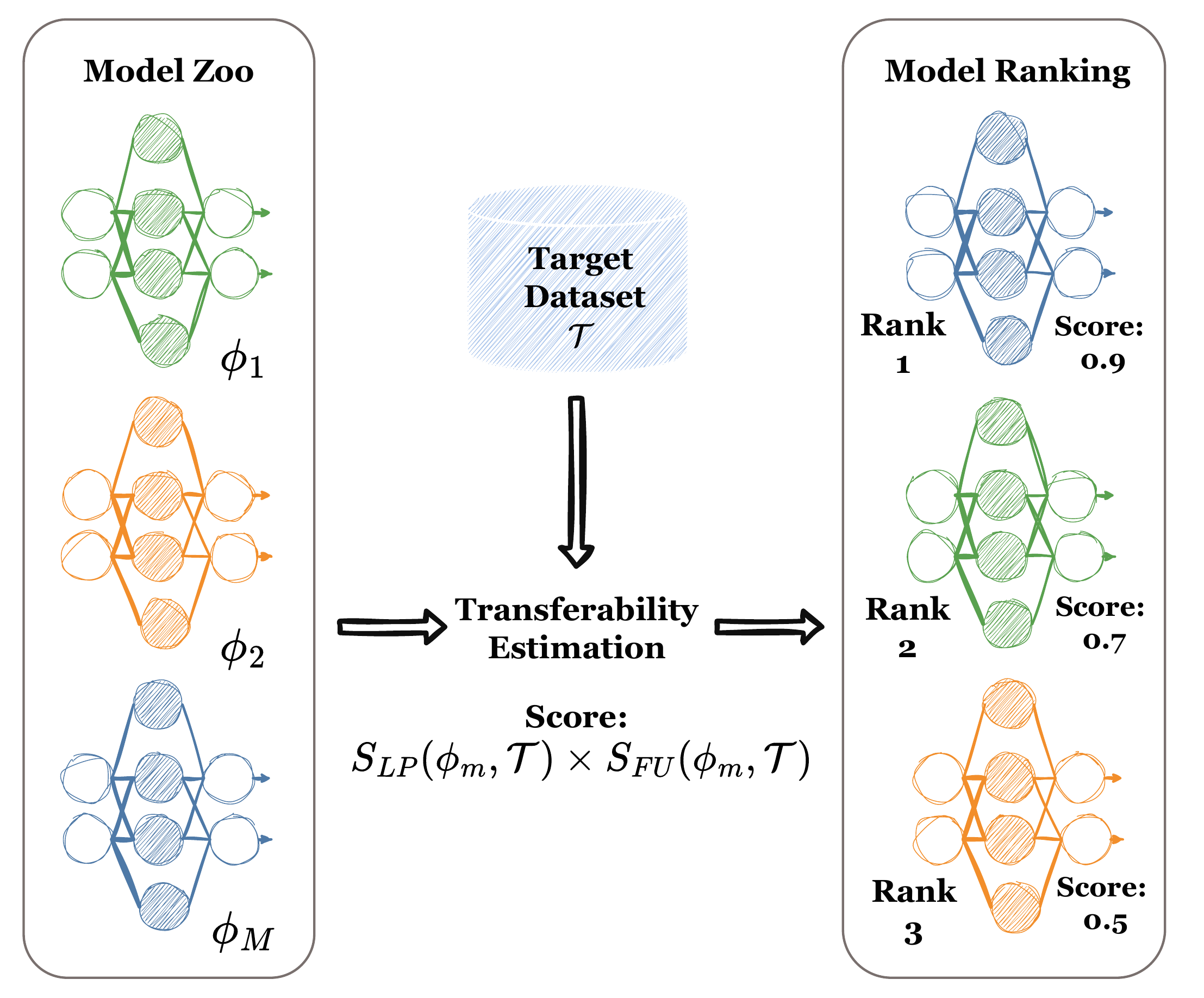} 
    \caption{Illustration of the transferability estimation problem: Given a model zoo, the goal is to predict which model will achieve higher performance after fine-tuning on a specific target task.}
    \label{fig:transferintro} 
\end{figure}
There is no reliable method to identify alternative datasets that might be better suited for transfer for medical image classification. Exhaustively fine-tuning multiple source models to determine suitability is computationally prohibitive. Transferability estimation in computer vision offers a solution by predicting how well pre-trained models will perform on new tasks or datasets without requiring extensive fine-tuning (Figure \ref{fig:transferintro}). This approach can efficiently uncover unexpected model candidates that human practitioners might otherwise overlook \cite{bassignana2022evidence}. As the number and complexity of pre-trained models grow, transferability estimation becomes increasingly valuable, enabling more effective reuse of source data and models.

The medical imaging community frequently repurposes models developed for general computer vision tasks for use in medical applications. However, as demonstrated by Chaves el al.~\cite{chaves2023performance} and further supported by experiments in this paper, current transferability metrics--designed and validated on natural image datasets--perform poorly when applied to medical image classification. This highlights the need to develop transferability metrics specifically tailored to medical imaging tasks.

Existing transferability methods primarily estimate the suitability of pre-trained features for a target task. However, feature quality alone is insufficient as it leads to self-source bias: if transferability were based solely on feature quality, a model pre-trained on the target dataset itself would provide the best features for that task. Yet, pre-trained models often outperform models trained exclusively on the target task, therefore it is likely that they also outperform models pre-trained directly on the target dataset. 
This is particularly relevant when models are pretrained on datasets much larger (and more varied) than those available for the target task.

We posit that transferability depends not only on the quality and generality of the pre-trained features but also on their flexibility, i.e., how easily new local patterns can be learned on the target task. We therefore propose a new transferability metric that balances both aspects
by incorporating the gradients of the first layers observed in the source model when exposed to the target dataset. The contributions of this paper are as follows:
\begin{itemize}
    \item We demonstrate that publicly available medical datasets, or combinations of them, can outperform ImageNet pre-training for medical image classification tasks.
    \item We establish two new testing scenarios to properly evaluate existing transferability metrics on medical imaging tasks: one for
    source dataset transfer in medical image classification and another for cross-domain transfer. We demonstrate that current state-of-the-art model selection methods fail to outperform simple baselines in these settings.
    \item We propose a novel transferability metric that combines feature quality with gradients, addressing the self-source bias of previous methods based solely on feature quality. We demonstrate that this method outperforms existing approaches.
    \item We provide ground-truth transfer performance benchmarking results for MedMNIST \cite{medmnistv1, medmnistv2}, a publicly available and easy-to-use benchmark dataset. This includes the transfer performance of 15 source datasets and 9 CNN architectures across 11 medical target tasks. We hope this will encourage further research in transferability estimation for medical image classification. To benchmark transfer performance, we trained over 20,000 models, a computationally intensive task that may otherwise discourage researchers with limited resources from exploring ideas in this field.
\end{itemize}

\section{Related work} \label{sec:related}
There are three main topics relevant to our work on transferability estation: dataset similarity, transferability metrics, and transferability specifically in medical imaging. 

\subsection{Dataset similarity}

Transferability estimation is closely related to dataset similarity, which can be measured using three main approaches: task similarity, embedding-based techniques, and distribution-based similarity estimation.

\paragraph{Task similarity} Transfer performance can serve as a proxy for task similarity, helping to reveal relationships between visual tasks. Zamir et al.~\cite{zamir2018taskonomy} mapped the structure of the space of visual tasks by computing transfer performance between pairs of tasks, creating an asymmetric similarity measure between source and target tasks that connects different tasks into a directed hypergraph, which is then pruned to produce a taxonomic map, or Taskonomy, of visual tasks. However, adding a new task to the Taskonomy is computationally expensive and it requires computing transfer performance on all previous tasks.

\paragraph{Embedding-based techniques} An alternative approach estimates dataset similarity directly to predict transfer performance, bypassing the need for fine-tuning, Embedding-based techniques establish a shared embedding space, where similarity is measured by the distance between task embeddings. Achille et al.~\cite{achille2019task2vec} employ a probe network trained on ImageNet \cite{deng2009imagenet} to vectorize tasks based on the Fisher information matrix of the network activations over a given dataset.  Similarly, Peng et al.~\cite{peng2020domain2vec} propose a domain embedding method that incorporates adversarially trained, domain-specific features. They compute the Gram matrix of activations from a pre-trained network over domain inputs, then concatenate its diagonal entries with domain-specific features extracted using a feature disentangler trained adversarially to separate domain-specific from class-specific features. These methods offer promising results but rely heavily on pre-trained probe models and still require model training on each dataset to some extent. 

\paragraph{Dataset distributions}
Dataset similarity can also be measured by directly comparing dataset distributions. For example the Optimal Transport Dataset Distance (OTDD) \cite{alvarez2020geometric} accounts for both sample and label distances by modeling labels as distributions over feature vectors and incorporating the Wasserstein distance between these distributions into the total transportation cost of the dataset samples. However, this approach require access to the source training set and overlook the impact of a model’s architecture, parameters, and training algorithms on transferability.

\subsection{Transferability metrics}

Transferability estimation methods can be broadly categorized into two main approaches: evaluating the quality of static features extracted by the source model when applied directly to the target dataset~\cite{nguyen2020leep,li2021ranking,pandy2022transferability,bolya2021scalable}, and modeling the changes that occur in the model during the fine-tuning process~\cite{you2021logme,shao2022not,wang2023far,li2023exploring}.

\paragraph{Static features}
The static features approach assumes that if the source model or its extracted features perform well on the target task, the knowledge encoded by the source model is valuable for the target task and is likely to transfer well.

Nguyen et al.~\cite{nguyen2020leep} proposed Log Expected Empirical Prediction (LEEP), which estimates the joint distribution between the source model's output labels and the target dataset labels. An empirical predictor is constructed from this distribution to capture the likelihood of target labels given the source predictions, and the LEEP score is derived as the log expectation of this predictor. However, LEEP inherently depends on the specific label space of the source model, limiting its applicability to models with classification heads. Gaussian LEEP ($\mathcal{N}$LEEP) \cite{li2021ranking} extends LEEP to support unsupervised and self-supervised pre-trained models that lack a classification head. By replacing the output layer with a Gaussian Mixture Model fitted to the target dataset in the source model’s penultimate embedding space, $\mathcal{N}$LEEP enables the computation of a LEEP score without relying on explicit source label probabilities.

Gaussian Bhattacharyya Coefficient (GBC) \cite{pandy2022transferability} introduces a different approach by measuring the pairwise class overlaps in distribution density with a Bhattacharyya coefficient, offering a versatile transferability metric applicable to image classification and semantic segmentation. Pairwise Annotation Representation Comparison (PARC) \cite{bolya2021scalable} evaluates transferability through Spearman correlation between the pairwise distance among target images in the feature space of the source model and the pairwise distance between the target labels. While these methods effectively measure feature quality, they may overlook the dynamic changes in representations that occur during fine-tuning, which can have a significant impact on transfer performance.

\paragraph{Modeling changes that occur during fine-tuning}
The second general approach accounts for changes that occur during fine-tuning, aiming to approximate this process and capture its effects on transfer performance.

The Logarithm of Maximum Evidence (LogME) \cite{you2021logme} adds a Bayesian linear model to the target features extracted by the source model and optimizes the parameters to estimate the likelihood of target labels given these features. Self-challenging Fisher Discriminant Analysis (SFDA) \cite{shao2022not} simulates the fine-tuning process by mapping features into a Fisher space to enhance between-class separability, while the self-challenging mechanism regularizes the model to focus on and improve differentiation of hard examples. 

NCTI \cite{wang2023far} builds on the concept of neural collapse \cite{papyan2020prevalence}, a phenomenon observed in the final stage of training, where features collapse to their class means and align in a structured geometric configuration. NCTI measures how far the source model is from this state on the target set by combining within-class variability, the simplex-encoded label interpolation geometry, and nearest-center classifier accuracy. Potential Energy Decline (PED) \cite{li2023exploring} is a physics-inspired approach that introduces a novel energy-based perspective, treating the fine-tuning process as a physical system minimizing potential energy. By modeling feature dynamics during adaptation, PED offers a feature remapping framework that can be integrated with existing methods for enhanced performance.

\paragraph{Our approach}
Our work falls into the second category as it models fine-tuning dynamics to evaluate transferability. However, we extend this approach by integrating gradient information to assess the adaptability of the source model's features to the target task. Existing methods tend to predict the target dataset as its own optimal source, a result that is not always realistic. We propose a transferability metric that avoids this self-source bias of prior methods.

\begin{figure*}[t]
\centering
\includegraphics[width=1.0\textwidth]{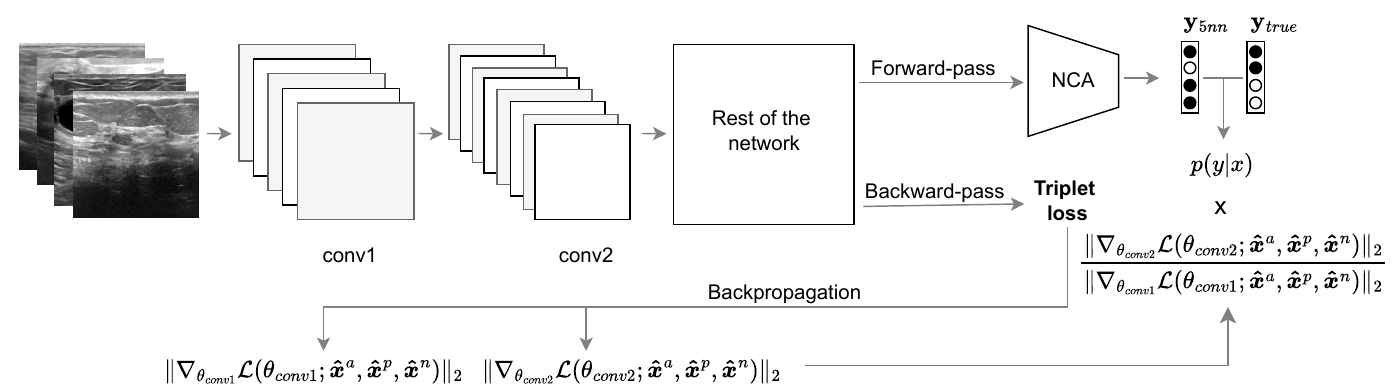}
\caption{Overview of our method. We use Neighborhood Component Analysis (NCA) on feature representations obtained from a forward pass of the target dataset to model fine-tuning dynamics and estimate the source model's feature suitability for the target task. It is then combined with the ratio of gradients from the second and first convolutional layers, obtained from the backward-pass, to estimate the magnitude of feature map updates in these layers during fine-tuning.}\label{method}
\end{figure*}

\subsection{Transferability in medical imaging}

Transferability metrics have primarily been developed and tested on natural image datasets, with limited exploration in the medical imaging domain. Chaves et al.~\cite{chaves2023performance} demonstrated that metrics designed for natural image datasets often fail to generalize to medical image classification tasks. Yang et al.~\cite{yang2023pick} proposed a method for medical image segmentation that combines class consistency and feature variety (CC-FV). The method measures intra-class consistency by calculating the distance between the distributions of features extracted from foreground voxels of the same class in each sample. Feature diversity is assessed by evaluating the uniformity of feature distribution across the entire global feature map, which reflects the effectiveness of the extracted features. Molina-Moreno et al.~\cite{molina2023automated} introduce a style-aware contrastive similarity estimator, trained to minimize a combined objective function that combines image reconstruction, style features, and dataset membership. Dataset similarity is then evaluated in the resulting latent space, which has been shown to correlate with transfer performance on target datasets, however, adding new source datasets requires retraining the estimator. Similar to the computer vision Taskonomy \cite{zamir2018taskonomy}, Du et al.~\cite{du2024datamap} build a DataMap of medical imaging datasets. Transferability is measured by the cosine similarity of task-relevant convolutional kernels from the last few convolutional layers of models trained on different datasets. However, while the DataMap captures symmetrical similarity between datasets, transferability is inherently asymmetrical--for instance, a larger and more diverse dataset is often a good source for a smaller dataset but not vice versa \cite{cheplygina2019cats,mensink2021factors}. We propose a transferability metric that is asymmetrical and does not require training.

\section{Method} \label{sec:method}

\subsection{Problem definition}

Given a set of models pre-trained on $M$ source datasets $\{\phi_1, \phi_2,..., \phi_M \}$, and a target dataset $\mathcal{T}=\{(\boldsymbol{x}_i,y_i)\}^n_{i=1}$ with $N$ labeled data points, the goal is to identify pre-trained models that are likely to perform well on the given target dataset, and to identify them \emph{without} requiring computationally expensive fine-tuning. 
The ground-truth transfer performance of a pre-trained model $\phi_m$ ($m \in \{1, 2, \dots, M\}$) when fine-tuned until convergence on $\mathcal{T}$ is measured using an evaluation metric $P(\phi_m,\mathcal{T})$, such as Area Under the Receiver Operating Characteristic Curve (AUC). Fine-tuning all $m$ models on $\mathcal{T}$ to compute $P(\phi,\mathcal{T})$ for each $\phi$ involves hyperparameter optimization and is computationally expensive, making it infeasible for large-scale source dataset selection.
The objective is to design a scoring function $S(\phi,\mathcal{T})$ for each pre-trained model $\phi_m$ such that the scores $S(\phi_m,\mathcal{T})$ correlate strongly with the true transfer performance $P(\phi_m,\mathcal{T})$. Specifically, the ranking of models by $S(\phi,\mathcal{T})$ should approximate the ranking by $P(\phi,\mathcal{T})$: 


\begin{equation}
\forall m \in \{1, \dots, M\}, \quad \rank(\{S(\phi_m,\mathcal{T})\}) \approx \rank(\{P(\phi_m,\mathcal{T})\}).
\end{equation}

\noindent Following prior work, we will measure this using weighted Kendall’s $\tau_w$ \cite{vigna2015weighted}, as it assigns greater importance to the correct ranking of top-performing models. A higher value of Kendall’s $\tau_w$ indicates a stronger correlation between $S(\phi,\mathcal{T})$ and $P(\phi,\mathcal{T})$.

\subsection{Gradient-based transferability estimation}

In contrast to previously proposed transferability metrics in computer vision that primarily focus on the suitability of pre-trained features for the target task, our method takes a more comprehensive approach, illustrated in Figure \ref{method}. Since medical targets appear to benefit less from feature reuse \cite{raghu2019transfusion, juodelyte2023revisiting}, we measure and incorporate the gradients of the first convolutional layers, computed from a single backward-pass on $\mathcal{T}$, to estimate their adaptation capabilities. We combine this with the feature representations of the target dataset $\mathcal{T}$ obtained from a single forward-pass through the pre-trained source model $\phi$.

\paragraph{Forward-pass}\label{forward-pass}
We use the penultimate layer of model $\phi_m$ to extract $D$-dimensional feature representation $\hat{\boldsymbol{x}} = \theta_m(\boldsymbol{x}) \in \mathbb{R}^D$ for each image $\boldsymbol{x}_i \in \mathcal{T}$ and use it as input to Neighborhood Component Analysis (NCA) \cite{goldberger2004neighbourhood} to approximate the dynamics of fine-tuning process.

\begin{figure}
\centering
\includegraphics[width=\linewidth]{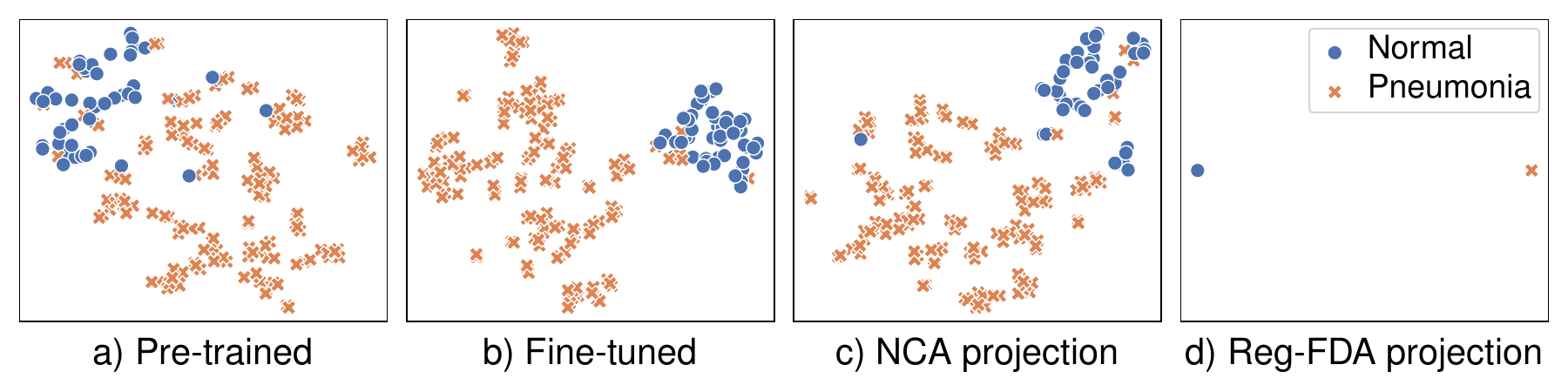}
\caption{t-SNE projections of feature representations $\hat{\boldsymbol{x}}$, for binary Pneumonia classification: (a) before fine-tuning the source model, (b) after fine-tuning, (c) after NCA projection, and (d) after LDA projection. The NCA projection (c) more closely approximates the fine-tuning dynamics, which update the features to achieve better class separability (b), compared to the LDA projection (d).}\label{projections}
\end{figure}

NCA is a supervised, non-parametric method used for dimensionality reduction and metric learning that directly learns a linear transformation to a lower-dimensional feature space where instances of the same class are clustered together, and instances of different classes are well-separated. The key idea in NCA is to maximize the probability that a randomly chosen point has the same label as its nearest neighbor in the transformed space. This is done by learning a linear transformation that maximizes the expected k-nearest neighbors (k-NN) leave-one-out classification accuracy on the training set. NCA reduces intra-class distances while increasing inter-class distances, effectively simulating the behavior of fine-tuning, which updates features to achieve better class separability \cite{shao2022not} (Figure \ref{projections}(b)). Once the projection matrix $\mathbf{A}$ is obtained, we compute updated feature representations $\{\tilde{\boldsymbol{x}}_i = \mathbf{A} \hat{\boldsymbol{x}}_i\}^n_{i=1}$. These transformed representations exhibit significantly improved class separability (as shown in Figure \ref{projections}(c)) compared to the original features before fine-tuning (Figure \ref{projections}(a)).

Shao et al.~\cite{shao2022not} use Fisher discriminant analysis (FDA) to approximate fine-tuning. FDA finds a linear subspace that maximizes class separability such that a linear classifier can be learned. However, FDA relies on the within-class scatter matrix, which requires the number of data points $n$ to far exceed the feature dimension $D$ ($n \gg D$).  If 
$D > n$, as is common in deep networks where $D$ is large (e.g., $D=
512$ for ResNet models), the sample covariance matrix becomes rank-deficient and cannot serve as a reliable estimator of the true covariance matrix \cite{park2022high}. Transfer learning scenarios often operate in low-data regimes where $n < D$, making FDA unsuitable. Shao et al.~\cite{shao2022not} use regularized FDA, which enhances robustness against outliers and numerical instability in scenarios with limited data points. However, as illustrated in Figure \ref{projections}(d), this approach results in all points from a class collapsing onto a single point in binary classification with $n=200$, indicating overfitting. In contrast, NCA avoids matrix inversion and does not enforce a linear decision boundary. Instead, it learns a robust transformation through a regularized linear projection \cite{goldberger2004neighbourhood}, providing a more reliable approximation of fine-tuning dynamics, as shown in Figure \ref{projections}(c).

Using the updated feature representations $\tilde{\boldsymbol{x}}_i$ we apply a 5-NN classifier to estimate the likelihood of target labels $p(y|\tilde{\boldsymbol{x}})$. The label prediction probability score $S_{LP}$ is then defined as:

\begin{equation}
S_{LP}(\phi_m,\mathcal{T}) = \sum^n_{i=1}  p(y_i|\tilde{\boldsymbol{x}}_i,\theta_m)
\end{equation}

\paragraph{Backward-pass}

The second component of our transferability relies on source model \textit{gradients}. As we do not have a classification layer, we compute the triplet loss using the feature representations $\hat{\boldsymbol{x}}$ obtained from the penultimate layer.
Triplet loss is widely used in deep metric learning tasks to learn a representation where similar samples are closer together in feature space, and dissimilar samples are farther apart. It operates on triplets of data points, where each triplet consists of:

\begin{itemize} \item Anchor ($\boldsymbol{x}^a$): a sample from a specific class. \item Positive ($\boldsymbol{x}^p$): another sample from the same class as the anchor. \item Negative ($\boldsymbol{x}^n$): a sample from a different class. \end{itemize}

\noindent We use the triplet margin loss \cite{wang2014learning}, implemented by \cite{musgrave2020PyTorchML}, with the goal of minimizing the distance between the anchor and the positive sample while ensuring that the distance between the anchor and the positive sample is at least $\alpha$ smaller than the distance between the anchor and the negative sample. The triplet margin loss is defined as:

\begin{eqnarray}
\mathcal{L}(\boldsymbol{\hat{x}}^a,\boldsymbol{\hat{x}}^p,\boldsymbol{\hat{x}}^n) = \max \{ \lVert \boldsymbol{\hat{x}}^a - \boldsymbol{\hat{x}}^p \rVert_2 - 
\lVert \boldsymbol{\hat{x}}^a - \boldsymbol{\hat{x}}^n \rVert_2 + \alpha, 0 \}
\end{eqnarray}

\noindent We perform a single backward pass of the triplet loss through the source model $\phi$ and compute the gradients w.r.t. the weights of the first two convolutional layers. Since gradients from models trained on different source datasets are not directly comparable, we calculate the \textit{ratio} of the gradient magnitudes of the second convolutional layer to the first convolutional layer. The first convolutional layer typically captures general features like edges and undergoes minimal updates (hence a good candidate for normalizing the size of the gradients), while deeper layers adapt more to the specific task. We hypothesize that this gradient ratio captures the model's task adaptability, i.e., its capacity to learn new local patterns. The feature update score $S_{FU}$ is defined as:

\begin{equation}
S_{FU}(\phi, \mathcal{T}) = \frac{\lVert \nabla_{\theta_{conv2}} \mathcal{L}(\theta_{conv2}; \boldsymbol{\hat{x}}^a,\boldsymbol{\hat{x}}^p,\boldsymbol{\hat{x}}^n) \rVert_2 }{\lVert \nabla_{\theta_{conv1}} \mathcal{L}(\theta_{conv1}; \boldsymbol{\hat{x}}^a,\boldsymbol{\hat{x}}^p,\boldsymbol{\hat{x}}^n) \rVert_2} 
\end{equation}

\noindent Both the label prediction probability score $S_{LP}(\phi_m, \mathcal{T})$ and the feature update score $S_{FU}(\phi_m, \mathcal{T})$ are normalized for consistency across tasks and datasets:

\begin{equation}
S_{FU}(\phi_m, \mathcal{T}) =  \frac{ S_{FU}(\phi_m, \mathcal{T}) - \min (S_{FU}(\phi, \mathcal{T}))}{ \max (S_{FU}(\phi, \mathcal{T})) - \min (S_{FU}(\phi, \mathcal{T}))} 
\end{equation}

\noindent The final transferability score is obtained as the sum of the normalized label prediction probability score and the normalized feature update score:

\begin{equation}
S(\phi_m, \mathcal{T}) = S_{LP}(\phi_m, \mathcal{T}) \times S_{FU}(\phi_m, \mathcal{T})
\end{equation}

\noindent This combined score effectively captures both the separability of the target features and the adaptability of the source model to new local patterns in the target task, providing a comprehensive transferability metric.

Finally, when selecting a source model $\phi_m$ for a given target task $\mathcal{T}$, we compute $S(\phi_m, \mathcal{T})$ for all pre-trained models and select the one with the highest score.

\begin{equation}
\phi^* = \argmax_{\phi_m} S(\phi_m, \mathcal{T}),
\end{equation}

\setlength{\tabcolsep}{5pt}
\begin{table}[t]
\centering
\resizebox{\linewidth}{!}{\begin{tabular}{lcrr}
\toprule
\textbf{Dataset} & \textbf{Modality} & \textbf{\# classes} & \textbf{batch size} \\
\midrule
PathMNIST \cite{kather2019predicting} & Colon pathology & 9 & $\{128, 256\}$\\ 
DermaMNIST \cite{tschandl2018ham10000, codella2019skin} & Dermatoscope & 7  & $\{128, 256\}$\\
OCTMNIST \cite{kermany2018identifying} & Retinal OCT & 4  & $\{64, 128\}$\\
PneumoniaMNIST \citep{kermany2018identifying} & Chest x-ray & 2 & $\{32, 64\}$\\
RetinaMNIST \cite{dataset20202nd} & Fundus ultrasound & 5  & $\{64, 128\}$\\
BreastMNIST \cite{al2020dataset} & Breast ultrasound & 2  & $\{32, 64\}$\\
BloodMNIST \cite{acevedo2020dataset} & Blood cell microscope & 8 & $\{128, 256\}$\\
TissueMNIST \cite{ljosa2012annotated} & Kidney cortex microscope & 8  & $\{128, 256\}$\\
OrganAMNIST \cite{xu2019efficient, bilic2023liver} & Abdominal CT & 11  & $\{128, 256\}$\\
OrganCMNIST \cite{xu2019efficient, bilic2023liver}  & Abdominal CT & 11  & $\{128, 256\}$\\
OrganSMNIST \cite{xu2019efficient, bilic2023liver}  & Abdominal CT & 11  & $\{128, 256\}$\\
\bottomrule
\end{tabular}}
\caption{Target datasets used in our experiments from the MedMNIST collection. Organ\{A,C,S\}MNIST are based on 3D computed tomography (CT) images, where A, C, and S represent Axial, Coronal, and Sagittal planes, respectively.}\label{targets}
\end{table}

\section{Experimental setup} \label{sec:setup}
In this section, we describe our experimental setup, choice of
datasets, models and hyperparameters.

\subsection{Datasets}
We evaluate our transferability metric using 11 out of the 12 datasets in the MedMNIST collection \cite{medmnistv1, medmnistv2} as target datasets 
$\mathcal{T}$. We exclude the Chest dataset (originally from  \cite{wang2017chestx}) as a target because it is a multi-label dataset with 14 classes, which is a less common scenario for transfer learning since smaller target datasets typically benefit more from pre-training. 

To simulate realistic transfer learning scenarios, we downsample each target dataset to include 100 images per class for training and 25 images per class for validation, both preserving class distributions. The ground-truth transfer performance $P(\phi_m, \mathcal{T})$ is evaluated using the full test sets. Table \ref{targets} provides a detailed overview of the target datasets, including their modality, the number of classes, and the batch sizes used during fine-tuning.

For source datasets, we use all 12 datasets in the MedMNIST collection, employing ResNet18 \cite{he2016deep} models trained on these datasets, as provided by MedMNIST. Additionally, we implement a leave-target-out pre-training strategy for MedMNIST datasets, wherein the target dataset is excluded from the pre-training set. For example, if Path is the target, all other datasets in MedMNIST are used for pre-training except Path.

Beyond MedMNIST, we include two additional source datasets for pre-training: ImageNet \cite{deng2009imagenet}, a large-scale natural image dataset widely used for pre-training, and RadImageNet \cite{mei2022radimagenet}, a specialized medical imaging dataset comprising CT, MRI, and ultrasound images.

\begin{figure*}[t]
\centering
\includegraphics[width=\textwidth]{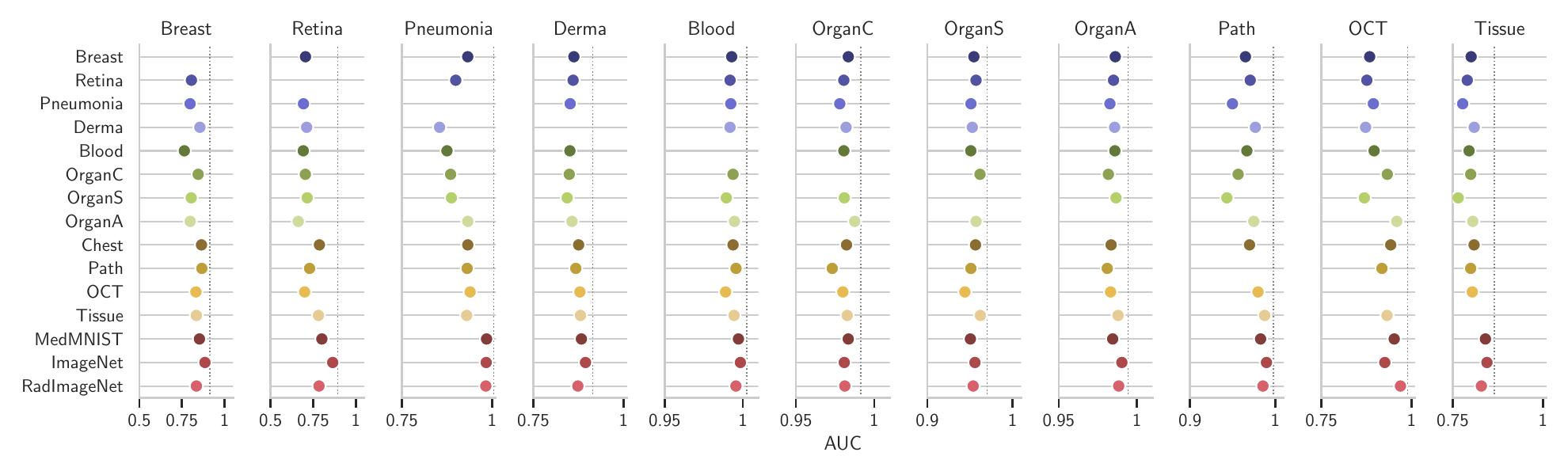}
\caption{Transfer performance (AUC) of source datasets (y-axis) evaluated on target test sets. Source datasets are sorted by size, from smallest to largest. The grey dashed line represents the best transfer performance for each target for easier comparison. Overall we do not see a relationship between source data size and transfer performance.}\label{AUC}
\end{figure*}

\subsection{Benchmarking transfer performance}

To benchmark ground-truth transfer performance $P(\phi_m, \mathcal{T})$, we perform full fine-tuning of the source models, with no weights frozen.  Hyperparameter optimization is conducted using a grid search over key hyperparameters \cite{li2020rethinking}, including the learning rate: $(\{1.0, 1e-01, 1e-02, 1e-03, 1e-04, 1e-05\})$, weight decay $(\{1e-03, 1e-04, 1e-05, 1e-06, 0.0\})$, momentum $(\{0.9, 0.0\})$, and batch size for each target dataset as specified in Table \ref{targets}. Hyperparameters and fine-tuning results are logged using radT \cite{robroek2023data}.

Transfer performance is measured using AUC. Training is carried out using the Stochastic Gradient Descent (SGD) optimizer for up to 400 epochs. To prevent overfitting, early stopping is applied if the validation AUC does not improve for 50 consecutive epochs. The model with the lowest validation loss is selected as the best-performing model, and its ground-truth transfer performance is evaluated by computing the AUC on the test set.

Input images are resized to $224 \times 224$ pixels and augmented using a series of transformations. These include random horizontal flips with a probability of 0.5, random resized cropping, random rotation between $0^{\circ}$ and $5^{\circ}$, random sharpness adjustment, random autocontrast, and random equalization. Images are normalized to the ImageNet mean and standard deviation when transferring from ImageNet-trained models. For models trained on other source datasets, images are normalized to have a mean of 0.5 and a standard deviation of 0.5.

\section{Results} \label{sec:results}
In this section we analyze the experimental results, highlighting important findings of our study.

\subsection{Transfer performance}

\textbf{Dataset size does not predict transfer performance.} 
In Figure \ref{AUC}, source datasets are sorted by size, from smallest to largest, to analyze the impact of dataset size on transfer performance. The results reveal no clear trend that would indicate that larger datasets inherently lead to better transfer performance. For example, the Breast dataset, containing only 546 training images, outperforms the much larger OrganS dataset, which has 13,932 training images, in 7 out of 9 target datasets (excluding Breast and OrganS as targets). Breast outperforms OrganS in Blood, Derma, OCT, Pneumonia, Path, and Tissue by a large margin. Even in cases where the target is related, such as OrganC and OrganA, which are different 2D planes of the same 3D dataset as OrganS, Breast performs comparably to OrganS, with AUC scores of 0.984 versus 0.981 for OrganC and 0.986 versus 0.987 for OrganA, respectively. This indicates that dataset size alone is not a reliable predictor of transferability performance.

\textbf{Similarity is not enough.} Similarity between source and target datasets does not necessarily result in optimal transfer performance. For instance, while both the Chest and Pneumonia datasets consist of chest X-rays, with Chest including a pneumonia class, the Chest does not achieve the best performance on the Pneumonia target. Instead, RadImageNet, ImageNet, and leave-target-out MedMNIST pre-training outperform it, achieving AUC scores of 98.24, 98.35, and 98.45, respectively. Notably, the leave-target-out MedMNIST dataset, which incorporates a variety of modalities (including chest X-rays), achieves the best performance. This suggests that, in addition to task-specific features, the diversity of data plays an important role in improving transfer learning performance by incorporating relevant yet distinct knowledge from the source dataset.

\textbf{Source dataset diversity is important.}
Leave-target-out MedMNIST outperforms RadImageNet in 7 out of 11 target datasets (Blood, Breast, Derma, OrganC, Pneumonia, Retina, and Tissue) despite being less than half its size. This may be attributed to the broader range of imaging modalities included in leave-target-out MedMNIST, compared to RadImageNet, which is limited to CT, MR, and ultrasound images with relatively low variation both within and between classes. Although leave-target-out MedMNIST has fewer classes (between 57 and 66, depending on the target) compared to RadImageNet (165 classes), the greater diversity in its training data appears to improve transfer learning performance.

\textbf{Medical sources outperform ImageNet in some cases.} While ImageNet pre-training remains a strong baseline, medical-specific source datasets outperform it in 4 out of 11 target tasks (Pneumonia, OrganC, OrganS, and OCT). For instance, in the OCT target, RadImageNet outperforms ImageNet by a large margin (AUC score of 96.93 versus 92.58). These results show that although ImageNet often performs well due to its large scale, diversity, and general features, exploring medical-specific source datasets can lead to improved performance for medical target tasks. 

\subsection{Transferability estimation}

\begin{figure*}[t!]
\centering
\includegraphics[width=0.88\textwidth]{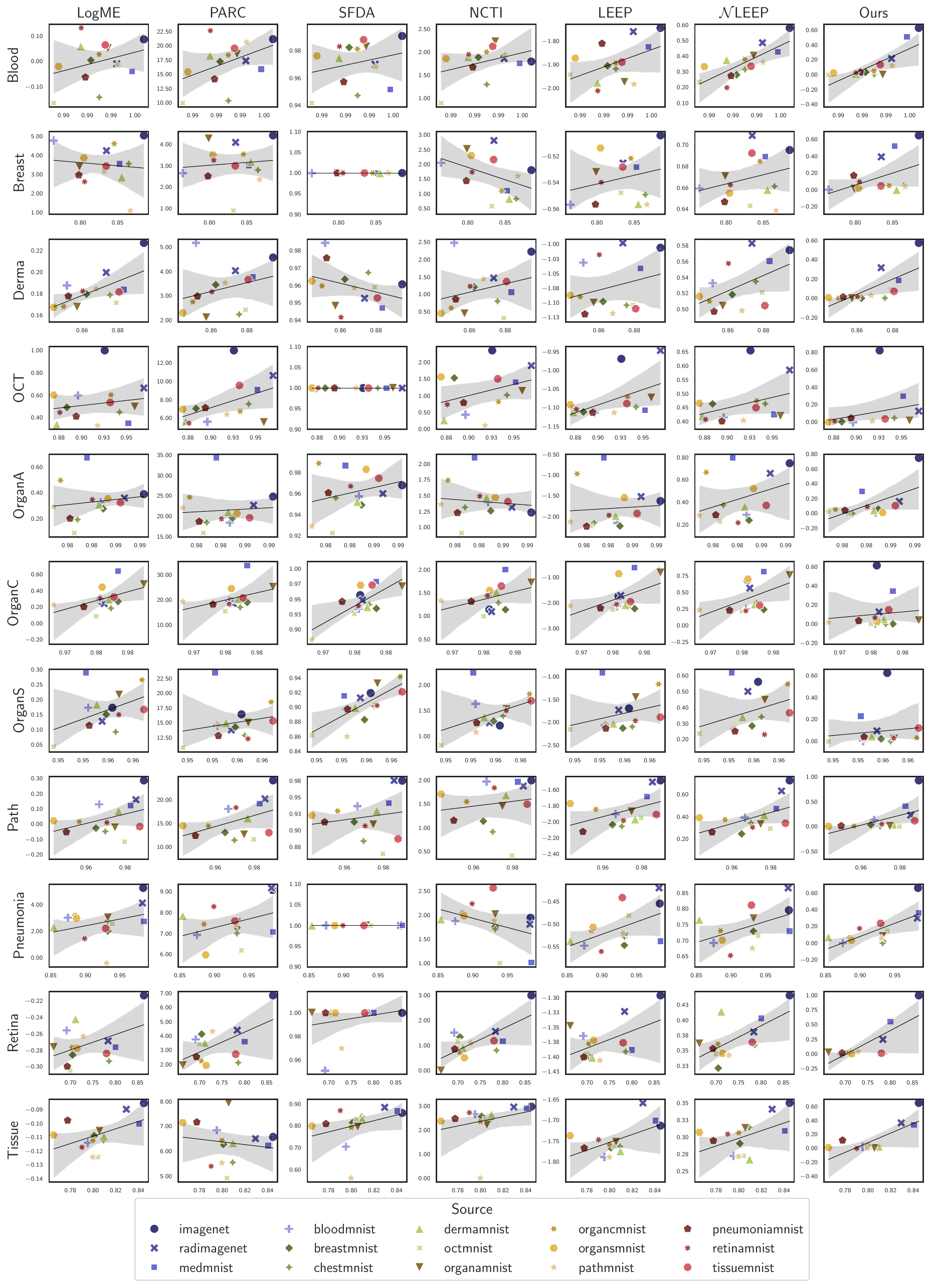}
\caption{Ground-truth transfer performance $P(\phi_m, \mathcal{T})$ (test AUC) on the x-axis versus transferability score $S(\phi_m, \mathcal{T})$ on the y-axis. The predicted transferability scores are shown for LogME, LEEP, SFDA, PARC, NCTI, $\mathcal{N}$LEEP, and our method (columns) across 11 medical target datasets (rows). The black line represents the regression line, with the 95\% confidence interval shaded in grey.}\label{result}
\end{figure*}

\setlength{\tabcolsep}{1pt}
\begin{table}[t]
\centering
\resizebox{\linewidth}{!}{\begin{tabularx}{0.6\textwidth}{lr *{6}{Y}}
\toprule
\textbf{Target} & \textbf{LogME} & \textbf{PARC} & \textbf{SFDA} & \textbf{NCTI} & \textbf{LEEP} & \textbf{$\mathcal{N}$LEEP} & \textbf{Ours} \\
\midrule
Blood & 0.11 (6)& 0.30 (4)& 0.30 (5)& 0.07 (7)& 0.48 (3)& \underline{0.75} (2)& \textbf{0.78} (1) \\
Breast & 0.22 (3)& 0.20 (5)& - (7)& -0.15 (6)& \underline{0.26} (2)& 0.21 (4)& \textbf{0.44} (1)\\
Derma & \underline{0.52} (2)& 0.34 (4)& -0.27 (7)& 0.19 (6)& 0.23 (5)& 0.44 (3)& \textbf{0.70} (1)\\
OCT & 0.26 (5)& 0.34 (3)& 0.19 (7)& 0.27 (4)& \textbf{0.52} (1)& 0.23 (6)& \underline{0.45} (2)\\
OrganA & 0.26 (4)& 0.27 (3)& -0.00 (6)& -0.26 (7)& 0.18 (5)& \underline{0.32} (2)& \textbf{0.57} (1)\\
OrganC & \underline{0.47} (2)& \textbf{0.50} (1)& 0.43 (5)& 0.46 (4)& 0.40 (6)& 0.47 (3)& 0.11 (7)\\
OrganS & 0.12 (6)& 0.17 (5)& \textbf{0.66} (1)& \underline{0.31} (2)& 0.24 (3)& 0.10 (7)& 0.22 (4)\\
Path & 0.51 (5)& 0.54 (4)& 0.38 (7)& 0.40 (6)& 0.56 (3)& \underline{0.57} (2)& \textbf{0.62} (1)\\
Pneumonia & 0.31 (3)& 0.12 (5)& - (7)& -0.37 (6)& 0.22 (4)& \underline{0.31} (2)& \textbf{0.61} (1)\\
Retina & 0.33 (6)& 0.50 (3)& 0.27 (7)& 0.47 (4)& 0.34 (5)& \underline{0.55} (2)& \textbf{0.60} (1)\\
Tissue & 0.58 (3)& -0.01 (7)& 0.46 (5)& \underline{0.62} (2)& 0.42 (6)& 0.58 (4)& \textbf{0.65} (1)\\
\midrule
Avg. rank & \textbf{4.00} & \textbf{4.00} & 5.82 & 4.91 & \textbf{3.91} & \textbf{3.45} & \textbf{1.91} \\
\bottomrule
\end{tabularx}}
\caption{Comparison of transferability metrics for dataset transferability prediction, evaluated using Weighted Kendall's $\tau$ between the predicted transferability scores and ground-truth transfer performance. Higher values indicate better performance, with the corresponding method rankings shown in parentheses (lower ranks are better). The best results are in bold, and the second-best results are underlined. The last row shows the average ranks. Statistical significance is determined by the Friedman test, with methods in bold indicating either the best performance or no significant difference from the best.}\label{wtau_data}
\end{table}

Transferability methods are typically tested in scenarios where models are pre-trained and fine-tuned on natural images. In contrast, we assess transferability metrics in two distinct scenarios: (1) we use multiple source datasets with a fixed architecture fine-tuned on medical targets to evaluate source dataset transferability estimation in medical imaging classification, and (2) we use multiple architectures pre-trained on ImageNet and fine-tuned on medical targets to evaluate model transferability estimation in a cross-domain transfer context. We benchmark our proposed transferability estimation method against existing methods, including LogME \cite{you2021logme}, SFDA \cite{shao2022not}, PARC \cite{bolya2021scalable},  NCTI \cite{wang2023far}, LEEP \cite{nguyen2020leep}, and $\mathcal{N}$LEEP \cite{li2021ranking}. 

\paragraph{Dataset transferability} 
We begin by fine-tuning ResNet18 \cite{he2016deep}, pre-trained on 14 source datasets, on 11 medical target datasets. The results of this experiment are presented in Figure \ref{result}. Our proposed method, along with LEEP and $\mathcal{N}$LEEP, are the only methods to consistently show a positive rank correlation with the ground-truth transfer performance $P(\phi_m, \mathcal{T})$  across all target datasets. In contrast, other methods exhibit negative correlations for at least one target. SFDA, in particular, struggles with binary classification tasks. On the Breast and Pneumonia targets, SFDA assigns a uniform transferability score of 1.0 to all source datasets, reflecting overfitting in the low-data regime, as discussed in Section \ref{forward-pass}. Even for the multiclass OCT target, SFDA lacks nuance, predicting a transferability score of 1.0 for nearly all source datasets, with only a minor adjustment to 0.99 for the Blood dataset. This lack of granularity severely limits SFDA’s utility in identifying suitable candidates for fine-tuning.

Table \ref{wtau_data} further highlights the strong performance of our proposed method, which achieves the highest rank correlation $\tau_w$ with the ground truth on eight target datasets and ranks second-best on one additional target. Notably, there is no single method that consistently performs well when our method does not, nor is there a clear runner-up that reliably ranks second. However, $\mathcal{N}$LEEP emerges as the overall second-best method. Our method outperforms $\mathcal{N}$LEEP by 0.40, 0.25, 0.24, 0.21, 0.20, and 0.12 rank correlation $\tau_w$ on Pneumonia, OrganA, Derma, OCT, Tissue, and OrganS, respectively. Despite these strengths, our method shows a substantial performance gap on the OCT, OrganC, and OrganS targets compared to the best-performing transferability metric for those datasets.

To evaluate whether the differences between the transferability methods are statistically significant, we use the Friedman test, as recommended by \cite{demvsar2006statistical}. Friedman test is a non-parametric statistical test designed for comparing ranks across multiple datasets. The average ranks of the methods are shown in the last row of Table \ref{wtau_data}. For SFDA, we assign the lowest rank for missing $\tau_w$ values on binary classification tasks, as its uniform prediction of a transferability score of 1.0 for all sources fails to provide useful guidance for selecting candidates for fine-tuning.

The Friedman test rejects the null hypothesis--that observed rank differences are due to chance--with a $p\text{-value} = 0.002$. Using a significance level of $\alpha = 0.05$, the critical difference (CD) for 11 datasets and seven methods is calculated as 2.792. Based on this CD, although our method achieves the highest average rank, the ranks of $\mathcal{N}$LEEP, LEEP, LogME, and PARC fall within the critical difference threshold of 2.792, indicating that their performance differences are not statistically significant. As the critical difference increases with the number of methods compared, and decreases with the number of the datasets, we expect that with experiments on additional datasets, the superior performance of our method would be more pronounced.

\begin{figure}[]
\centering
\includegraphics[width=0.5\textwidth]{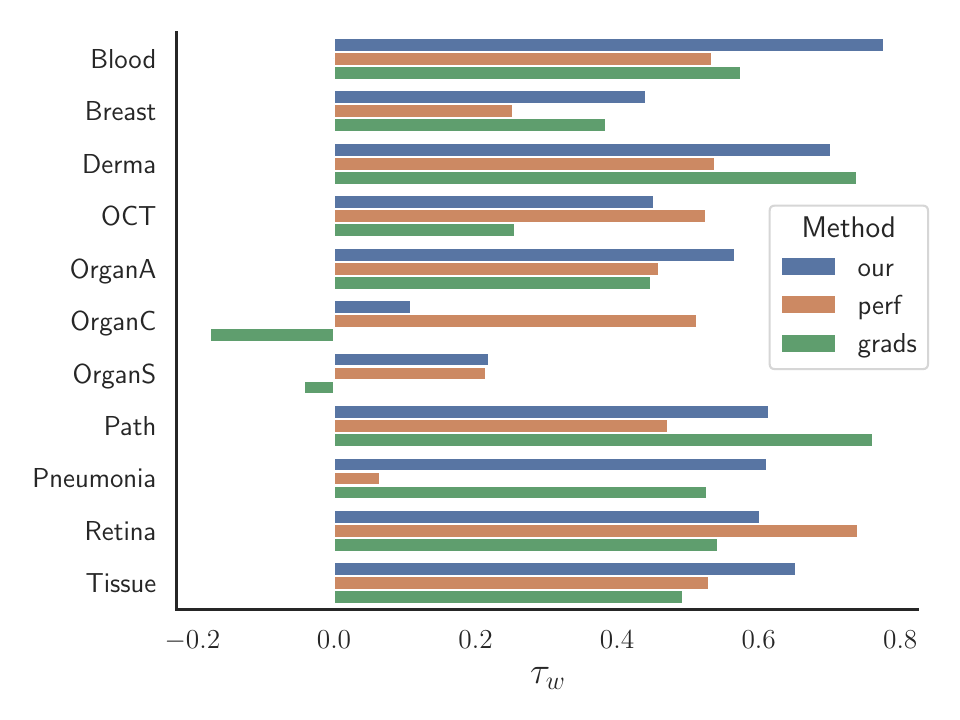}
\caption{Contribution of the feature quality $S_{LP}(\phi_m, \mathcal{T})$ and feature update $S_{FU}(\phi_m, \mathcal{T})$ terms to the overall transferability score.}\label{ablation}
\end{figure}

Interestingly, in this source selection experiment, simpler empirical conditional probability-based methods, such as LEEP and $\mathcal{N}$LEEP--among the earliest proposed transferability metrics--outperform more recent, sophisticated methods like NCTI and SFDA that explicitly model the feature space.

\paragraph{Ablation study} Our proposed metric is composed of two terms: one evaluates the suitability of the features for the target task, and the other estimates the feature update during fine-tuning. To assess the contribution of each component, we conduct an ablation study, with the results presented in Figure \ref{ablation}. The analysis reveals that the feature update term often enhances the overall transferability estimation, particularly in targets such as Blood, Breast, Derma, OrganA, Path, Pneumonia, and Tissue. This underscores the value of incorporating gradient-based information into transferability metrics. However, for targets where our method underperforms, the feature update term has a negative impact, suggesting the need for refining how feature updates are estimated. 

Additionally, the feature quality term alone outperforms LogME, SFDA, and NCTI in seven targets. This result highlights the effectiveness of modeling feature space using NCA in our approach, which appears to capture transfer dynamics more effectively than Reg-FDA, LDA, PCA, or the linear models used in those methods.

\setlength{\tabcolsep}{1pt}
\begin{table}[]
\centering
\resizebox{0.485\textwidth}{!}{\begin{tabularx}{0.75\textwidth}{lr *{6}{Y}}
\toprule
\textbf{Target} & \textbf{LogME} & \textbf{1-PARC} & \textbf{SFDA} & \textbf{1-NCTI} & \textbf{1-LEEP} & \textbf{1-$\mathcal{N}$LEEP} & \textbf{1-Ours} \\
\midrule
Blood & -0.20 (7)& -0.01 (6)& 0.28 (4)& 0.32 (3)& \textbf{0.48} (1)& \underline{0.35} (2)& -0.00 (5)\\
Breast & -0.23 (7)& -0.18 (6)& \underline{0.40} (2)& \textbf{0.69} (1)& 0.22 (3)& 0.17 (5)& 0.22 (4)\\
Derma & -0.15 (7)& \textbf{0.46} (1)& 0.23 (4)& -0.01 (6)& \underline{0.45} (2)& 0.38 (3)& 0.19 (5)\\
OCT & 0.26 (5)& \underline{0.51} (2)& 0.16 (6)& 0.38 (4)& 0.44 (3)& \textbf{0.66} (1)& 0.11 (7)\\
OrganA & 0.36 (5)& \underline{0.60} (2)& 0.33 (6)& -0.00 (7)& 0.53 (3)& 0.52 (4)& \textbf{0.70} (1)\\
OrganC & \textbf{0.29} (1)& 0.09 (5)& 0.12 (3)& 0.01 (7)& 0.10 (4)& 0.04 (6)& \underline{0.19} (2)\\
OrganS & \underline{0.39} (2)& 0.21 (4)& 0.04 (5)& -0.14 (7)& 0.35 (3)& -0.06 (6)& \textbf{0.42} (1)\\
Path & -0.50 (5)& -0.75 (7)& \underline{0.09} (2)& -0.39 (4)& \textbf{0.14} (1)& -0.51 (6)& -0.30 (3)\\
Pneumonia & -0.08 (7)& 0.28 (3)& 0.16 (5)& \underline{0.29} (2)& 0.25 (4)& 0.12 (6)& \textbf{0.32} (1)\\
Retina & 0.29 (3)& -0.18 (7)& \underline{0.40} (2)& 0.09 (4)& -0.09 (5)& -0.12 (6)& \textbf{0.70} (1)\\
Tissue & -0.03 (5)& \underline{0.21} (2)& -0.15 (7)& -0.15 (6)& 0.19 (3)& -0.01 (4)& \textbf{0.32} (1)\\
\midrule
Avg. rank & \textbf{4.91} & \textbf{4.09} & \textbf{3.91} & \textbf{4.64} & \textbf{2.91} & \textbf{4.73} & \textbf{2.82} \\
\bottomrule
\end{tabularx}
}
\caption{Comparison of transferability metrics for model transferability prediction, evaluated using Weighted Kendall's $\tau$ between the predicted transferability scores and ground-truth transfer performance. Higher values indicate better performance, with the corresponding method rankings shown in parentheses (lower ranks are better). The best results are in bold, and the second-best results are underlined. The last row shows the average ranks. Statistical significance, determined by the Friedman test, indicates no significant difference between the methods.}\label{wtau_imnet}
\end{table}

\begin{table*}[]
\centering
\resizebox{\textwidth}{!}{\begin{tabularx}{1.2\textwidth}{lr *{10}{Y}}
\toprule
\textbf{Source} & \textbf{Blood} & \textbf{Breast} & \textbf{Derma} & \textbf{OCT} & \textbf{OrganA} & \textbf{OrganC} & \textbf{OrganS} & \textbf{Path} & \textbf{Pneumonia} & \textbf{Retina} & \textbf{Tissue} \\
\midrule
ImageNet & 99.85 & 88.51 & 89.46 & 92.58 & 99.04 & 98.09 & 95.58 & 98.97 & 98.35 & 86.41 & 84.49 \\
RadImageNet & 99.56 & 83.50 & 87.35 & 96.93 & 98.84 & 98.12 & 95.38 & 98.55 & 98.24 & 78.39 & 82.94 \\
MedMNIST & 99.72 & 85.28 & 88.34 & 95.19 & 98.45 & 98.34 & 95.05 & 98.28 & 98.45 & 80.04 & 84.06 \\
Blood &  & 76.42 & 85.14 & 89.63 & 98.59 & 98.06 & 95.10 & 96.66 & 87.48 & 69.15 & 79.53 \\
Breast & 99.29 & & 86.25 & 88.38 & 98.62 & 98.35 & 95.46 & 96.50 & 93.23 & 70.44 & 80.10 \\
Chest & 99.38 & 86.49 & 87.57 & 94.22 & 98.35 & 98.23 & 95.65 & 96.98 & 93.25 & 78.61 & 80.93 \\
Derma & 99.18 & 85.55 &  & 87.25 & 98.58 & 98.20 & 95.27 & 97.66 & 85.42 & 71.12 & 80.96 \\
OCT & 98.90 & 83.21 & 87.92 &  & 98.32 & 98.00 & 94.40 & 97.97 & 93.89 & 69.93 & 80.42 \\
OrganA & 99.47 & 79.89 & 85.71 & 95.91 & & 98.76 & 95.72 & 97.48 & 93.25 & 66.21 & 80.57 \\
OrganC & 99.37 & 84.54 & 84.96 & 93.28 & 98.18 &  & 96.18 & 95.64 & 88.49 & 70.35 & 79.99 \\
OrganS & 98.95 & 80.47 & 84.40 & 86.91 & 98.67 & 98.09 &  & 94.32 & 88.73 & 71.45 & 76.52 \\
Path & 99.57 & 86.70 & 86.78 & 91.79 & 98.11 & 97.32 & 95.11 &  & 93.07 & 72.85 & 79.98 \\
Pneumonia & 99.23 & 79.80 & 85.22 & 89.42 & 98.28 & 97.80 & 95.12 & 94.98 &  & 69.27 & 77.76 \\
Retina & 99.19 & 80.58 & 86.03 & 87.60 & 98.51 & 98.06 & 95.71 & 97.06 & 89.90 &  & 79.02 \\
Tissue & 99.44 & 83.46 & 88.07 & 93.20 & 98.80 & 98.28 & 96.21 & 98.75 & 92.94 & 78.05 &  \\
\bottomrule
\end{tabularx}}
\caption{Ground-truth transfer performance (test set AUC $\times 100$) of source datasets across various medical targets.}\label{medmnist}
\end{table*}

\begin{table*}[t]
\centering
\resizebox{\textwidth}{!}{\begin{tabularx}{1.2\textwidth}{lr *{10}{Y}}
\toprule
\textbf{Source} & \textbf{Blood} & \textbf{Breast} & \textbf{Derma} & \textbf{OCT} & \textbf{OrganA} & \textbf{OrganC} & \textbf{OrganS} & \textbf{Path} & \textbf{Pneumonia} & \textbf{Retina} & \textbf{Tissue} \\
\midrule
DenseNet & 99.90 & 90.14 & 87.70 & 98.18 & 99.05 & 98.64 & 95.51 & 99.02 & 98.73 & 87.81 & 84.34 \\
EfficientNet & 99.92 & 88.60 & 91.15 & 99.03 & 99.29 & 98.56 & 95.60 & 98.74 & 97.94 & 87.22 & 84.21 \\
GoogleNet & 99.77 & 88.18 & 88.78 & 96.44 & 99.15 & 98.39 & 95.64 & 99.37 & 98.94 & 83.42 & 84.21 \\
MnasNet & 99.53 & 85.86 & 77.09 & 93.79 & 95.88 & 94.13 & 90.06 & 98.78 & 89.21 & 80.85 & 78.31 \\
MobileNet & 99.82 & 88.72 & 89.44 & 94.64 & 98.86 & 98.74 & 94.39 & 99.30 & 97.90 & 84.21 & 83.69 \\
VGG & 99.69 & 86.34 & 90.30 & 98.03 & 99.30 & 98.93 & 96.44 & 98.88 & 98.43 & 87.73 & 85.09 \\
ConvNeXt & 99.91 & 91.58 & 92.93 & 98.86 & 99.27 & 98.87 & 96.38 & 99.47 & 97.97 & 87.55 & 85.30 \\
ShuffleNet & 99.74 & 84.82 & 89.15 & 97.17 & 98.94 & 98.65 & 96.02 & 99.37 & 98.05 & 80.52 & 83.65 \\
ResNet & 99.85 & 88.51 & 89.46 & 92.58 & 99.04 & 98.09 & 95.58 & 98.97 & 98.35 & 86.41 & 84.49 \\
\bottomrule
\end{tabularx}}
\caption{Ground-truth transfer performance (test set AUC $\times 100$) of CNN architectures pre-trained on ImageNet across various medical targets.}\label{imnet}
\end{table*}

\paragraph{Model transferability in cross-domain transfer}
We further assess the transferability estimation metrics on ranking pre-trained CNN models. For this evaluation, we select nine widely-used architectures: ResNet18 \cite{he2016deep}, DenseNet121 \cite{huang2017densely}, EfficientNetV2-S \cite{tan2021efficientnetv2}, MobileNetV3-Small \cite{howard2019searching}, GoogleNet \cite{szegedy2015going}, MnasNet-1.0 \cite{tan2019mnasnet}, VGG11 \cite{simonyan2014very}, ConvNeXt-Tiny \cite{liu2022convnet}, and ShuffleNetV2-0.5x \cite{ma2018shufflenet}. All these models are pre-trained on ImageNet and available in PyTorch \cite{pytorch}. The results of this experiment are presented in Table \ref{wtau_imnet}.

In this scenario, none of the evaluated transferability metrics demonstrate a positive rank correlation across all target datasets. In fact, PARC, NCTI, LEEP, $\mathcal{N}$LEEP, and our proposed method predominantly have negative rank correlations. To address this, we transform the predictions of these methods to $1-S(\phi_m, \mathcal{T})$. For our method specifically, we normalize the feature quality and feature update terms before combining them, as follows:

\begin{equation}
S_{FU}(\phi_m, \mathcal{T}) =  \frac{ S_{FU}(\phi_m, \mathcal{T}) - \max (S_{FU}(\phi, \mathcal{T}))}{ \min (S_{FU}(\phi, \mathcal{T})) - \max (S_{FU}(\phi, \mathcal{T}))}
\end{equation}

This adjustment results in a positive rank correlation between the predicted transferability scores and the ground-truth transfer performance for the majority of targets. However, this contradicts intuition. For methods like PARC, NCTI, LEEP, and $\mathcal{N}$LEEP, the transformation implies that greater deviation between the predictions based on pre-trained features and the true labels corresponds to better transferability. Similarly, for our metric, the transformation suggests that smaller feature updates and greater prediction deviations correlate with improved transfer performance. These findings highlight a potential gap in our understanding of cross-domain transfer dynamics. The results suggest that knowledge transfer in cross-domain settings, especially for medical targets, may operate differently compared to in-domain transfer, emphasizing the need for rethinking how transferability is modeled in such contexts.

Our method outperforms other models on five target datasets. For targets where our method's transferability score predictions are negatively correlated with the ground-truth transfer performance, the difference in performance between the best- and worst-performing source models is minimal, with AUC differences of only 0.004 for Blood and 0.007 for Path. Although our method achieves the highest average rank, the Friedman test fails to reject the null hypothesis, indicating that the observed rank differences may be due to chance.

We encourage the research community to leverage the ground-truth transfer performance results provided in Tables \ref{medmnist} and \ref{imnet}, along with the resources available in our GitHub repository, to further explore this relatively underexplored topic. Transferability estimation not only holds significant practical potential but also offers opportunities to deepen our understanding of transfer learning in general. Our code is publicly available and has been designed to be extendable, facilitating the evaluation and integration of additional transferability metrics.

\section{Conclusions} \label{sec:conclusion}
This study proposed a novel transferability estimation measure for transfer learning in medical image classification, balancing both the suitability of the learned features for the target task and the model's adaptability, i.e., its capacity to learn new local patterns linked to subtle local texture variations.

We introduce a novel NCA-based transferability metric, the first to combine feature quality with gradient information from the first convolutional layers from a single backward-pass. We also propose two new testing scenarios for transferability estimation in medical imaging: one focused on source dataset transferability in medical image classification and the other on cross-domain transferability. The results show that our metric outperforms state-of-the-art methods which focus solely on feature suitability for the target task, such as SFDA and NCTI, in both scenarios. This highlights the importance of incorporating gradient information into transferability estimation.

We show that, while ImageNet remains a strong baseline, medical-specific source datasets outperform ImageNet in several medical target tasks. This underscores the value of selecting alternative source datasets for medical image classification tasks, which may offer benefits over relying on ImageNet as the default source dataset for pre-training.

Our experiments, spanning a diverse range of medical image classification tasks, reveal three key insights: (1) dataset size alone does not reliably predict transfer performance, (2) similarity between source and target datasets is not always sufficient for optimal transfer, and (3) the diversity of the source dataset plays a pivotal role in transfer performance.

Our results also suggest that a source model's feature suitability and adaptability may have a negative correlation with transfer performance in cross-domain transfer in some cases, particularly when transferring from natural to medical images. This highlights a gap in our understanding of cross-domain transfer dynamics. Indeed, knowledge transfer in cross-domain settings fundamentally differs from in-domain transfer, and may require more elaborate source model selection and fine-tuning techniques. To support further research in this underexplored field, we also provide detailed transfer performance data for 15 source datasets, 9 model architectures, and 11 target datasets, including hyperparameter optimization, encompassing over 20,000 trained models. 

This work paves the way for future advancements in transferability estimation methods, offering a promising foundation for improving medical image classification and empowering more accurate, reliable healthcare solutions.

\vspace{10pt}
\noindent \textbf{Acknowledgments.} This study was funded by the Novo Nordisk Foundation (grant number NNF21OC0068816). 


\bibliographystyle{elsarticle-num} 
\bibliography{refs_manual}



\end{document}